\documentclass{article}

\pdfoutput=1

    \usepackage[preprint]{neurips_2020}

\usepackage[utf8]{inputenc} 
\usepackage[T1]{fontenc}    
\usepackage{hyperref}       
\usepackage{url}            
\usepackage{booktabs}       
\usepackage{amsfonts}       
\usepackage{nicefrac}       
\usepackage{microtype}      
\usepackage{graphicx}
\usepackage{amsmath}
\usepackage{amsthm}
\usepackage{amssymb}
\usepackage{natbib}
\setcitestyle{square}
\usepackage[labelfont=bf]{caption}
\graphicspath{ {./}}
\usepackage{float}
\title{A priori guarantees of finite-time convergence for Deep Neural Networks}
\author{
  Anushree Rankawat   \thanks{Currently at Universit\'e de Montr\'eal. Work was done when the author was at Ahmedabad University. \newline
  Correspondence to: Anushree Rankawat  \href{anushree.rankawa@gmail.com}{<anushree.rankawat@gmail.com>}}\\
  Ahmedabad University\\
  \texttt{anushree.rankawat@gmail.com} \\
  \And
  Mansi Rankawat \\
  \texttt{mansirankawat19@gmail.com} \\ 
  \And
  Harshal B. Oza \\
   Pandit Deendayal Petroleum University\\
   \texttt{harshal.b.oza@gmail.com} \\
}

\newcommand{\sign}{\mathrm{sign}}

\begin{document}

\maketitle

\begin{abstract}
  In this paper, we perform Lyapunov based analysis of the loss function to derive an a priori upper bound on the settling time of deep neural networks. While previous studies have attempted to understand deep learning using control theory framework, there is limited work on a priori finite time convergence analysis. Drawing from the advances in analysis of finite-time control of non-linear systems, we provide a priori guarantees of finite-time convergence in a deterministic control theoretic setting. We formulate the supervised learning framework as a control problem where weights of the network are control inputs and learning translates into a tracking problem. An analytical formula for finite-time upper bound on settling time is computed a priori under the assumptions of boundedness of input. Finally, we prove the robustness and sensitivity of the loss function against input perturbations. 
\end{abstract}

\section{Introduction}
Deep neural networks have seen significant developments over the past decade, with some achieving human-like performance in various machine learning tasks, such as classification, natural language processing and speech recognition. Despite the popularity of deep learning, the underlying theoretical understanding remains relatively less explored. While attempts have been made to develop deep learning theory by drawing inspiration from other related fields such as statistical learning and information theory, a comprehensive theoretical framework remains in a developing stage. Deep neural networks have been proven to perform well with an increase in the depth of the network, which consequently results in an increase in the number of parameters. Though this drastically improves the performance of the neural network, it ends up making the neural network less interpretable. The neural network can also become less conducive to mathematical analysis with increase in complexity. Other problems in deep neural networks revolve around the stability and desired convergence of the training. Since the performance of the network depends highly on the training data and the choice of the optimization algorithm, it may make the output of the network diverge for small perturbations. Our work attempts to give finite-time convergence guarantees for training of a deep neural network by utilizing an established stabilization framework from control theory.

Existing works in deep learning theory have attempted to bridge the gap in understanding deep learning dynamics by focusing on simple models of neural networks (\citet{saxe2013exact}, \citet{li2017convergence}, \citet{arora2018convergence}, \citet{jacot2018neural}). This could be attributed to the fact that current state-of-the-art deep learning models are highly complex structures to analyze. \citet{jacot2018neural} proves that a multilayer fully-connected network with infinite width converges to a deterministic limit at initialization and the rate of change of weights goes to zero. \citet{saxe2013exact} analyzed deep linear networks and proved that these networks, surprisingly, have a rich non-linear structure. The study shows that given the right initial conditions, deep linear networks are a finite amount slower than shallow networks. Following this work, \citet{arora2018convergence} proves the convergence of gradient descent to global minima for networks with dimensions of every layer being full rank in dimensions. While these studies give important insights into the design of neural network architecture and the behavior of training, these results may need to be modified for the conventional deep neural networks for the convergence guarantees. \citet{du2018gradient} extends the work of \citet{jacot2018neural} further by proving convergence for gradient descent to achieve zero training loss in deep neural networks with residual connections. 

When it comes to convergence of certain state variables of a dynamical system, control theory provides a rich mathematical framework which can be utilized for analyzing the non-linear dynamics of deep learning(\citet{liu2019deep}). One of the early works relating deep learning to control theory was of \citet{lecun1988theoretical}, which used the concept of optimal control and formulated back-propagation as an optimization problem with non-linear constraints. Non-linear control has gained increasing attention over the past few years in the context of neural networks, especially for recurrent neural networks (\citet{allen2019convergence}, \citet{xiao2017accelerating}) and reinforcement learning (\citet{xu2013reinforcement}, \citet{gupta2019finite}, \citet{wang2019multistep}, \citet{kaledin2020finite}). A new class of recurrent neural networks, called Zhang Neural Networks (ZNN), was developed that expressed dynamics of the network as a set of ordinary differential equations and used non-linear control to prove global or exponential stability for time-varying Sylvester equation (\citet{zhang2002recurrent}, \citet{guo2011zhang}). \citet{li2013accelerating} introduces the sign bi-power activation function for Zhang Neural Networks (ZNN) which helps the author prove the existence of finite-time convergence property. \citet{haber2017stable} presents deep learning as a parameter estimation problem of non-linear dynamical systems to tackle the exploding and vanishing gradients.

The focus of this paper is on deriving a priori guarantee of attaining finite-time convergence of training under some assumptions on inputs. A supervised learning framework is considered with arbitrary number of hidden layers. The novelty lies in the fact that the loss function is treated as a Lyapunov function. The temporal derivative of Lyapunov function is ensured to be negative definite in that it follows a particular differential inequality. The resulting training update is derived naturally as a function of time such that it ensures the convergence of Lyapunov function to zero in finite time. The weight update as a function of time is in fact defined as the control input to the neural network. The only major assumption used is that some inputs will have their magnitudes with a lower bound greater than zero. Thus the learning problem is converted into a finite time stabilization problem as studied rigorously in \cite{Bhat2000}. Our contributions are twofold. First, the training problem of deep learning is converted into a tracking control problem under some reasonable assumptions. Second, the synthesis of the weight update as an input to the non-linear control system, the neural network, is carried out using the concepts of finite-time stabilization. To the best of our knowledge, a guarantee of finite-time convergence is being studied for the first time for training of a general multi-layer neural network. The proposed results will enable time bound training that may be useful in real-time applications and open new opportunities in viewing the learning problem from a deterministic systems viewpoint. 

The paper is organized as follows. Section \ref{s:ProposedAlgo} starts with introducing the loss function as a candidate Lyapunov function for a single neuron case and proves that this simple discriminant satisfies the conditions required for finite-time stability theorems developed in \cite{Bhat2000} to be applicable. Section \ref{s:MLP} then proves that a similar result extends to a multi-layer perceptron network under reasonable assumptions on the input to the layer in question. Section \ref{s:MLP} also provides an extension to the case when bounded perturbations are admitted at the input in that convergence guarantees are shown to hold true. In Section \ref{s:Experiments}, some numerical simulations are presented for both single neuron and multi-layer perceptron cases for regression. Section \ref{s:conclusion} collects conclusions and discusses future scope. 


\section{Proposed algorithm for weight update}\label{s:ProposedAlgo}
This section motivates the development of a priori bounds on settling time with certain assumptions admitted for the input. The weight update problem for supervised learning in neural networks is similar to the tracking problem of non-linear control systems. It is interesting to explore how traditional control theory extends to the realm of neural networks. We study a loss function which is similar to the usual loss functions that we encounter in machine learning. The rationale of our paper lies in defining the loss function as a candidate Lyapunov function. In turn, standard arguments of Lyapunov theorems lead to the conclusion that the mathematical problem of training neural networks can admit valid Lyapunov functions.  

Section \ref{s:SingleNeuron} develops the proof of convergence for the single neuron case. This simple case motivates an extension of the method to encompass loss functions for the analysis of multi-layer case. 

\subsection{Motivation}\label{s:SingleNeuron}
We start with a simplistic single neuron network. Let \(x \in \mathbb{R}^{n}\) define the input to the network where $x=\begin{bmatrix}x_1 & x_2 & \cdots & x_n\end{bmatrix}^\top$, $|x_i|<a, i=1,2,\cdots n$ holds true for some a priori but arbitrary scalar $a\in(0,\infty)$. Since we are dealing with supervised learning, we are also provided with the target output $y^\star$. The linear combination of weights $w_i$ with inputs $x_i$ is represented by $z= \sum_{i=1}^{n} w_ix_i+b$, where $b$ is a bias. As is usually done in defining the output behavior, $z$ is passed through a non-linear activation function, $\sigma(z)$. The following sigmoid function is chosen for defining the activation function for the present case: 

\begin{equation}\label{e:Activation}
    \sigma(z)= \frac{1}{1+e^{-z}}
\end{equation}

The output of the neural network is given by $y= \sigma(z)$. The first main concept in this paper is to treat the loss function as the candidate Lypunov function. First, let the error in output be defined as $\bar{e}=y-y^*$. Next, consider a continuous function $E(\bar{e})$ to be a candidate Lyapunov function as follows: 
 \begin{equation}\label{e:LossLyapunov}
        E = \frac{|\bar{e}|^{(\alpha+1)}}{(\alpha+1)},
 \end{equation}
where $\alpha\in(0,1)$ is a scalar to be chosen by the user. The second main concept in this paper is to define the temporal rate of weight as the control input to enforce the stability of the origin $\bar{e}=0$ as $t\rightarrow \infty$. The Lyapunov function \eqref{e:LossLyapunov} is used to show that it is indeed plausible to achieve this asymptotic stability goal. Taking the temporal derivative of the candidate Lyapunov function \eqref{e:LossLyapunov} produces
 
\begin{equation}
    \frac{\mathrm{d}E}{\mathrm{d}t} = \frac{\mathrm{d}E}{\mathrm{d}\bar{e}}\frac{\mathrm{d}\bar{e}}{\mathrm{d}y}\frac{\mathrm{d}y}{\mathrm{d}z}\frac{\mathrm{d}z}{\mathrm{d}w}\frac{\mathrm{d}w}{\mathrm{d}t}, \nonumber
\end{equation}
which can be computed using \eqref{e:Activation} through \eqref{e:LossLyapunov} as
\begin{equation}\label{e:Lyapderivativ}
    \frac{dE}{dt} = |\bar{e}|^{\alpha}\sign(\bar{e})\left(\frac{e^{-z}}{(1+e^{-z})^2}\right)\left(x_1\dot{w}_1 + x_2\dot{w}_2 + \cdots + x_n\dot{w}_n\right) 
\end{equation}
Define, $u_1\triangleq \dot{w}_1, u_2\triangleq \dot{w}_2, \cdots, u_n\triangleq \dot{w}_n$ and 
\begin{equation}\label{e:Control}
	u_i = -k_i\sign(x_i)\sign(\bar{e})e^{z}(1+e^{-z})^2, \quad i=1,2,\cdots,n,
\end{equation}    
where $k_i>0$, for all $i$, are tuning parameters to be chosen by the user. It can be noted that all control inputs $u_1, u_2, \cdots, u_n$ remain bounded due to boundedness assumption of all the inputs $x_i$ and that of $e^{z}$. Substituting \eqref{e:Control} into \eqref{e:Lyapderivativ} produces
\begin{equation}\label{e:Lyapderivativ1}
	\frac{dE}{dt} = -|\bar{e}|^{\alpha}\left(\sum\limits_{i=1}^{n}k_i|x_i|\right).
\end{equation}
\newtheorem{assum1}{Assumption}
\begin{assum1}\label{assum:1} 
	At least one input of all $x_i, i=1,2,\cdots,n$ is non-zero such that $|x_j|>\gamma>0$ where $\gamma$ is a priori known scalar for some integers $j\in[1,n]$.
\end{assum1}
It can be noted that Assumption \ref{assum:1} is not unreasonable for most practical applications in that some inputs will always be nonzero with a known lower bound on its magnitude.
First main result of the paper is in order.
\newtheorem{thm1}{Theorem}
\begin{thm1}\label{th:tm1h}
	Let assumption \ref{assum:1} hold true and let the output of the neural network be given by $y= \sigma(z)$. Let all the inputs $x_i, i=1,2,\cdots, n$ be bounded by some a priori known scalar $a\in(0,\infty)$ such that $|x_i|<a$ holds true for all $i$. Then, weight update \eqref{e:Control} causes the error $\bar{e}=y-y^*$ to converge to zero in finite time.
\end{thm1}    
\begin{proof}
	The proof of the theorem is furnished using standard Lyapunov analysis arguments. Consider $E$ defined by \eqref{e:LossLyapunov} as a candidate Lyapunov function. Observing \eqref{e:Lyapderivativ1}, it can be concluded that the right hand side of the temporal derivative of remains negative definite since it involves square terms and norm of inputs. Furthermore, \eqref{e:Lyapderivativ1} can be rewritten under the assumption \ref{assum:1} as follows:
	\begin{equation}\label{e:Lyapderivativ2}
		\frac{dE}{dt} \leq - k_{\mathrm{min}} \gamma E^{\beta},
	\end{equation}
	where $k_{\mathrm{min}} = \min(k_i), i=1,2,\cdots,n$, $\beta = \frac{\alpha}{\alpha+1}$ and $|\bar{e}|^{\alpha}=\left(|\bar{e}|^{\alpha+1}\right)^{\frac{\alpha}{\alpha+1}}=E^\beta$ has been utilized. Noting that $E$ is a positive definite function and scalars $k_{\mathrm{min}}$ ad $\gamma$ are always positive, the proof is complete by applying \citep[Theorem~4.2]{Bhat2000}.
\end{proof}

Theorem \ref{th:tm1h} provides a finite settling-time of our loss function. The upper bound on the settling time is available from \citep[Theorem~4.2]{Bhat2000}. 


\subsection{Multi Neuron Case}\label{s:MLP}
Consider a multi-layer perceptron with $N$ layers where the layers are connected in a feed-forward manner \citep[Chapter 4]{book:Bishop}. Let \(x \in \mathbb{R}^{n}\) define the input to the network where $x=\begin{bmatrix}x_1 & x_2 & \cdots & x_n\end{bmatrix}^\top$ and $|x_i|<a, i=1,2,\cdots n$ holds true for some a priori but arbitrary scalar $a\in(0,\infty)$. Let \(y \in \mathbb{R}^{m}\) define the multi-neuron output to the network where \(y=\begin{bmatrix}y_1 & y_2 & \cdots & y_m\end{bmatrix}\). Let \(y^{\star} \in \mathbb{R}^{m}\) define the target output values to the network where \(y^\star=\begin{bmatrix}y^{\star}_1 & y^{\star}_2 & \cdots & y^{\star}_m\end{bmatrix}\). The error in the output layer can be expressed as \(\bar{e} = \begin{bmatrix} |y_1-y^\star_1| & |y_2-y^\star_2| & \cdots & |y_m-y^\star_m|\end{bmatrix}\). Hence, the scalar candidate Lyapunov function can be written as follows:


\begin{equation}
        E = E_1+E_2+ \cdots + E_m
 \end{equation}
\begin{equation}\label{e:MultiLayerLoss}
        E =  \frac{|\bar{e}_1|^{(\alpha+1)}}{(\alpha+1)} + \frac{|\bar{e}_2|^{(\alpha+1)}}{(\alpha+1)} + \cdots \frac{|\bar{e}_m|^{(\alpha+1)}}{(\alpha+1)}
 \end{equation}
As is usually done in the case of feed-forward networks, consider unit $j$ of layer $l$ that computes its output 
\begin{equation}\label{e:UnitJ}
        a_j =  \sum\limits_{i} w_{ji}z_i,
\end{equation}
using its inputs $z_i$ from layer $l$ where bias parameter has been embedded inside the linear combination and $z_j=\sigma(a_j)$, where $\sigma$ is non-linear activation function. The aim of this section is to extend the single neuron case to a multi-neuron one. The simplest way do achieve this is to find sensitivity of $E$ to the weight $w_{ji}$ of a given layer $l$, which is given by
\begin{equation}\label{e:deltaj}
        \frac{\partial E_m}{\partial w_{ji}} = \frac{\partial E_m}{\partial a_j} \frac{\partial a_j}{\partial w_{ji}}.
\end{equation}
Using standard notation $\delta_j \triangleq  \frac{\partial E}{\partial a_j^l}$ and using \eqref{e:UnitJ}, the following results:
\begin{equation}\label{e:deltaj1}
        \frac{\partial E_m}{\partial w_{ji}} =  \delta_j z_i.
\end{equation}
It is straightforward to compute $\delta_m$ that belongs to the output layer as follows:
\begin{equation}\label{e:delta_op}
        \delta_m = \frac{\partial E_m}{\partial a_m} =  \sigma'(a_m) \frac{\partial E_m}{\partial y_m},
\end{equation}
where $z_m$ is replaced by $y_m$ as it is the output layer. Finally, computation of $\delta_j$ for all hidden units is given by
\begin{equation}\label{e:deltaj_hidden}
        \delta_j = \frac{\partial E_m}{\partial a_j} =  \sum\limits_{k} \frac{\partial E_m}{\partial a_k} \frac{\partial a_k}{\partial a_j},
\end{equation}
where units with a label $k$ includes either hidden layer units or an output unit in layer $l+1$. Combining \eqref{e:UnitJ}, $z_j=\sigma(a_j)$ and $\delta_j \triangleq  \frac{\partial E}{\partial a_j}$ produces
\begin{equation}\label{e:deltaj_hiddenfinal}
        \delta_j = \sigma'(a_j) \sum\limits_{k} w_{kj} \delta_k.
\end{equation}
A slightly modified version of assumption \ref{assum:1} is required before the next main result of the paper is now presented.
\newtheorem{assum2}[assum1]{Assumption}
\begin{assum2}\label{assum:2} 
	At least one input of all $z_i, i=1,2,\cdots,L$ is non-zero such that $|z_n|>\gamma>0$ where $\gamma$ is a priori known scalar for some integers $n\in[1,L]$ where $L$ is the number of units in layer $l$.
\end{assum2}
\newtheorem{thm2}[thm1]{Theorem}
\begin{thm2}\label{thm:2}
Let the weight update for connecting unit $i$ of layer $l$ to unit $j$ of layer $l+1$ of a multi-layer neural network be given by 
    \begin{equation}\label{e:weight_multi}
        \dot{w}_{ji} = -k_{ji}\sign(\delta_j z_i) |\delta_j z_i|^{\alpha}E^{\beta},
    \end{equation}
with some scalar $\beta\in(0,1)$ such that $\alpha+\beta<1$ and $k_{ji}>0$ is a tuning parameter. Then the output vector $y$ converges to $y^*$ in finite time. 
\end{thm2}
\begin{proof}
Consider the candidate Lyapunov function $E$ given by \eqref{e:MultiLayerLoss}. The temporal derivative of the Lyapunov function is given by 
\begin{equation}\label{e:LyapDerivativeMulti}
    \dot E = \sum\limits_{m} \dot{E}_m = \sum\limits_{m} \frac{\partial E_m}{\partial w_{ji}} \dot{w}_{ji},
\end{equation}
which can be simplified using \eqref{e:weight_multi} and \eqref{e:deltaj1} as 
\begin{equation}\label{e:LyapDerivativeMulti1}
    \dot E = - E^{\beta} \sum\limits_{m} k_{ji} | \delta_j z_i|^{\alpha+1}.
\end{equation}
Using Assumption \ref{assum:2} it is easy to conclude that for some $k_{\mathrm{min}}=\min\limits_{i,j,l}{k^l_{ji}}>0$, the following inequality holds true:
\begin{equation}\label{e:LyapDerivativeMulti2}
    \dot E \leq - k_{\mathrm{min}}\gamma^{\alpha+1} E^{\beta}. 
\end{equation}
Noting that $E$ is a positive definite function and scalars $k_{\mathrm{min}}$ ad $\gamma$ are always positive, the proof is complete by applying \citep[Theorem~4.2]{Bhat2000}.
\end{proof}

\newtheorem{rem1}{Remark}
\begin{rem1}
It can be seen that weight update \eqref{e:weight_multi} (or respectively \eqref{e:Control}) is in a feedback control form where state $z_i$ and $\delta_j$ (respectively $x_i$ and $\bar{e}$) are being used for influencing the learning process.
\end{rem1}

\subsection{Sensitivity to perturbations}
This section considers the robustness of training of the network when inputs deviate from their nominal values. The motivation to study such a scenario stems from the need to make the learning phase of a neural network less dependent on trial and error. If a neural network converges in the training phase favourably, then it is intuitive to expect it to converge when small perturbations are considered in the test data which might deviate by a known amount from the training data. This intuitive result is proven mathematically in this section. A priori guarantees were obtained in the previous section when lower and upper bounds on the inputs are admitted. This section develops theoretical claims where the same finite settling time as that derived in the case of unperturbed inputs is proven to hold true in the case of perturbed inputs. Hence, the training problem of a deep neural network is converted into a robust control problem. The following assumption of the upper bound on the perturbation of inputs is invoked.
\newtheorem{assum3}[assum1]{Assumption}
\begin{assum3}\label{assum:3} 
There exists an a priori known constant $M>0$ such that all inputs $x_i, i=1,2,\cdots,N$ admit additive perturbations $\Delta x_i$ such that 
	\begin{equation}\label{e:PertubatioBound}
	    |\Delta x_i| \leq M |x_i|^{\alpha},
	\end{equation}
for all $i$ where $N$ is the number of inputs.	
\end{assum3}
The following result is in order.
\newtheorem{thm3}[thm1]{Theorem}
\begin{thm3}
Let assumptions \ref{assum:2} and \ref{assum:3} hold true. Let the weight update for connecting unit $i$ of layer $l$ to unit $j$ of layer $l+1$ of a multi-layer neural network be given by \eqref{e:weight_multi}. Then the output vector $y$ converges to $y^*$ in finite time in the presence of additive perturbations $\Delta x_n, n\in[1,L]$ if $k_{\mathrm{min}}>M$. 
\end{thm3}
\begin{proof}
It can be seen that the perturbation is considered only in inputs. Hence all hidden layer weights are updated as done in the proof of Theorem \ref{thm:2}. Hence, the dynamics of learning results in the following revised temporal derivative of Lyapunov function:
\begin{equation}\label{e:LyapDerivativeMulti3}
    \dot E = - E^{\beta} \sum\limits_{m} k_{ji}| \delta_j z_i|^{\alpha+1} + \sum\limits_{k=0}^{n}  k_{1k} |\delta_kx_k|^{\alpha} \sign(\delta_k x_k)\delta_k\Delta x_k,
\end{equation}
where $k_{1k}$ is the gain parameter for training of all the input layer weights. The expression \eqref{e:LyapDerivativeMulti3} can be simplified using Assumption \ref{assum:3} as follows:
\begin{equation}\label{e:LyapDerivativeMulti4}
    \begin{aligned}
    \dot E &\leq - E^{\beta} \sum\limits_{m} k_{ji}|\delta_j z_i|^{\alpha+1} + E^{\beta}\sum\limits_{k=0}^{n}  k_{1k}|\delta_kx_k|^{\alpha+1}  M,\\
           & \leq - E^{\beta} \sum\limits_{m} (k_{ji} - M) | \delta_j z_i|^{\alpha+1},
    \end{aligned}
\end{equation}
where inputs $z_i$ now collect inputs $x_i$ as well. Similar to the proof of Theorem \ref{thm:2}, \eqref{e:LyapDerivativeMulti4} can be re-written by applying Assumption \ref{assum:2} as follows:
\begin{equation}\label{e:LyapDerivativeMultiPerturb2}
    \dot E \leq - (k_{\mathrm{min}}-M) \gamma E^{\beta}. 
\end{equation}
Since $k_{\mathrm{min}}>M$, the proof is complete by applying \citep[Theorem~4.2]{Bhat2000}.
\end{proof}

\section{Experiments}\label{s:Experiments}
This section presents empirical evidence for the results of the previous section thereby highlighting the importance of the proofs presented above. We observe that the Lyapunov Loss function converges more aggressively than the $L_2$ loss function. This is attributed mainly to the non-Lipschitz weight updates given by \eqref{e:Control} and \eqref{e:weight_multi}. Also, the time taken by the proposed loss function falls within the a priori upper bound. We also analyze the stability of convergence with respect to input perturbations.

For this, we have conducted two experiments. The first experiment pertains to the more illustrative example of single neuron case presented in Section \ref{s:SingleNeuron}, where we perform classification on the Iris dataset (\cite{Dua:2019}). For the last experiment, we consider the multi-layer perceptron case covered in Section \ref{s:MLP}, where we perform regression on the Boston Housing dataset (\citet{harrison1978hedonic}).

\textbf{Note.} By test loss at epoch $l$, we mean the loss we obtain when we feed a particular test example after the network has trained for $l$ epochs. This gives us a good representation of how the network has progressed with respect to predicting inputs it hasn't encountered yet.

\textbf{Experimental Setup.} For the above stated experiments, we have coded in Python on 4-core CPU with 32GB of RAM for training. All networks are trained on data with an 80-20 train data-test data split. The number of epochs for which the network is trained with the values of hyper-parameters is described for each case in their respective plots.

\begin{figure}[H]{
    \includegraphics[scale=0.23]{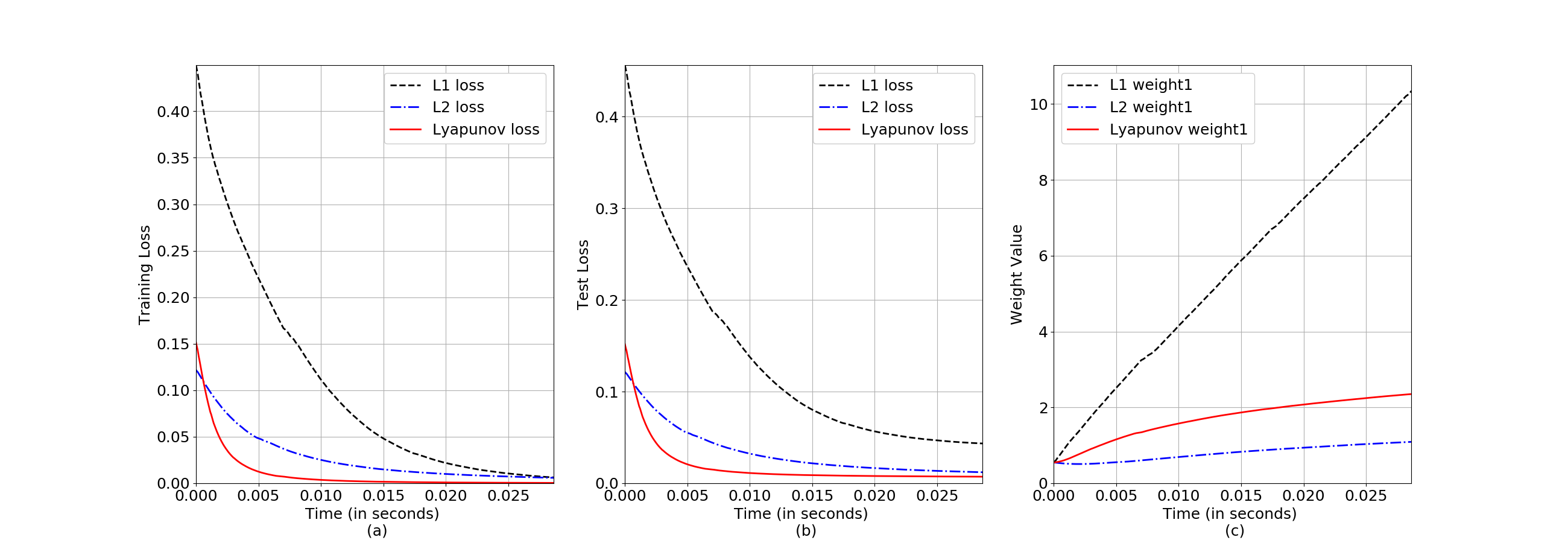}
    \centering
    \caption{Comparison of convergence with respect to time for a single neuron trained on the Iris dataset. We convert the problem into a binary classification problem by only considering inputs of two of the three classes. All networks are trained for 2100 epochs with 80 training examples and 20 test examples.} 
    \label{figure1}}
\end{figure}

\begin{figure}[H]{
    \includegraphics[scale=0.23]{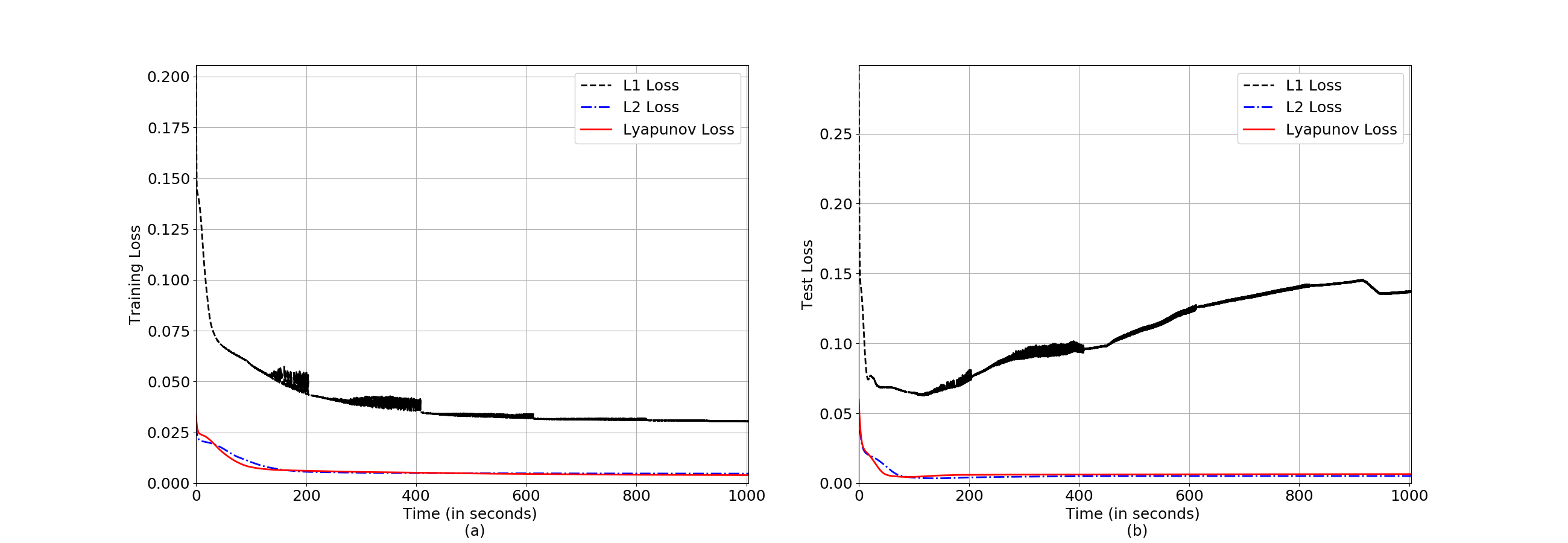}
    \centering
    \caption{Comparison of convergence with respect to time for the multi-layer perceptron trained on the Boston Housing dataset. All networks are trained for 40,000 epochs to observe the convergence.} 
    \label{figure2}}
\end{figure}

\textbf{Convergence Rate for training.} We consider the tasks of regression for single neuron and multi-layer perceptron networks. To demonstrate the improvement in convergence rate for the proposed Lyapunov Loss Function when compared to the traditional $L_1$ and $L_2$ loss functions, we  plot the corresponding training loss with respect to time in Figure \ref{figure1} and \ref{figure2}. The quantitative measure of when the network converges is reported in Table \ref{settling_time_table}. We observe that the proposed Lyapunov loss function always converges faster than $L_1$ and $L_2$, demonstrating the results proven extensively in Section \ref{s:ProposedAlgo}. Admittedly, the theoretical upper bound on settling time for single neuron and multi-layer perceptron produced by (\ref{e:Lyapderivativ2}) and (\ref{e:LyapDerivativeMulti2}) are very conservative, yet it is an a priori deterministic upper bound and the aggressive learning proves to be faster than traditional loss functions in the scenarios considered here. This can be attributed to the fact that we operate in the realm of non-smooth functions which tend to be more aggressive than the smooth $L_2$ function we usually encounter in most learning problems. While we can clearly observe that the loss converges to zero in finite-time for the single neuron case, the multi-neuron case settles at a value that happens to not be zero. This is because the current control framework is not equipped to handle constant bias at the current stage as only disturbances with a vanishing bound \eqref{e:PertubatioBound} are allowed. Of course, setting $\alpha=0$ can reject all persisting disturbances \cite{Orlov2005}, but this may result in discontinuity in back propagation and may result into large numerical errors. 

\begin{table}[ht]
  \centering
  \begin{tabular}{cccccc}
    \toprule
    \textbf{Experiment}  & \textbf{Theoretical Upper Bound}  & \multicolumn{3}{c}{\textbf{Experimental Convergence Time}}\\
     & (in seconds) & \multicolumn{3}{c}{(in seconds)} \\
     & & $L_1$ & $L_2$ & Lyapunov \\
     & & & & Loss function \\
    \hline
    Single Neuron (Iris) & $\sim$20224.17    & 0.0384 & 0.0388 & 0.0155\\ 
    MLP (Boston Housing) & $\sim$2.87*$e^{10}$ & 155.4587 & 179.2681 & 144.9205  \\
    \bottomrule
  \end{tabular}
  \caption{Settling time in seconds for each experiment conducted. We compare the time taken for convergence by three different loss functions, $L_1$, $L_2$ and Lyapunov Loss function. The training conditions were similar for individual cases in the experiment.}
  \label{settling_time_table}
\end{table}
\textbf{Hyperparameter Analysis and its effect on training.} We will discuss the impact of different values of $\alpha$ on the training process of our neural network. For the given experiments, we tested the network for various values of $\alpha$ ranging from 0-1. We observed that the training loss follows the proved equations as long as $\alpha \in (0.5,0.9)$. If $\alpha$ happens to be too close to zero, the resulting control law becomes discontinuous thereby resulting in numerical instabilities owing to the fact that the current ODE solvers are unable to handle functions that are discontinuous.

\begin{figure}[H]{
    \includegraphics[scale=0.5]{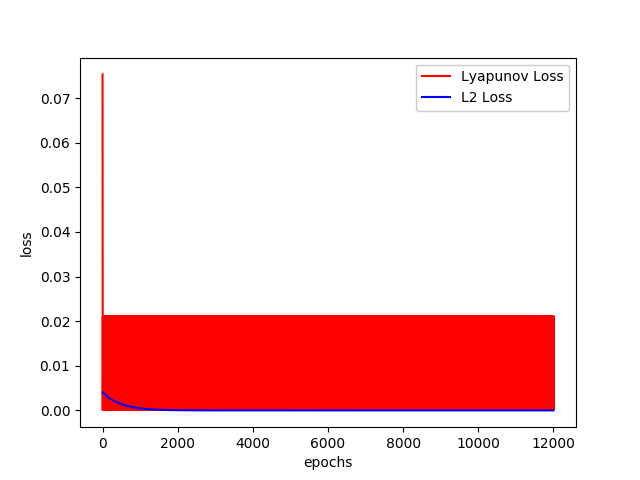}
    \centering
    \caption{Depiction of the numerical instability observed when we take $\alpha=0$. Effectively, at this point, the loss function contains a discontinuous signum function.} 
    \label{figure3}}
\end{figure}

\section{Conclusion and Future Work}\label{s:conclusion}
This paper studies the training of a deep neural network from the control theoretic viewpoint. In particular, the convergence of output error to a desired value is posed as a tracking control problem where weights of the network were seen as control inputs to the dynamical system. The developments naturally make use of the existing gradient methods along with an additional feedback term that accelerates the convergence of tracking error. The results are substantiated by existing results on Lyapunov analysis of finite time stable systems. The main theoretical contribution is to treat a non-quadratic loss function as the Lpyapunov function while synthesizing time rate of weight as a feedback law which is a non-Lipschtiz function of gradient of the loss function. Resulting performance is more aggressive in terms of convergence due to a non-Lipschitz nature of control update.

A good future scope is to convert the analysis framework from that which depends on continuous time to a more practical one that depends on epochs. This will make the presented results more accessible to a wider classes of applications. From a control theoretic viewpoint, such a conversion is rather standard, for example, when going from continuous time to a discrete time formulation. Also, the converse problem of finding the tuning parameters $k_{\mathrm{min}}, \alpha$ and $\beta$ for given settling time can also be a very useful result. 

\section*{Broader Impact}
The results presented in this paper aim for a deterministic approach towards training a neural network. A priori guarantee in supervised learning can pave the way for new frameworks that run in real time with faster convergence. The presented algorithm can be useful in time-critical applications, such as those employed with real time operating systems. Of course, a learning framework that converges to a decision in an apriori known time can have positive impact in areas such as healthcare. However, the impact of false negatives in a more aggressive learning such as that provided by the presented developments need to be analyzed more deeply.         

\bibliography{ms}
\bibliographystyle{plainnat}
\end{document}


\title

\textbf{Experimental Setup.} Since we deal with a more complicated multi-layer perceptron case which is trained on a large public dataset here, we use the NVIDIA GEFORCE GTX1080Ti GPU card with 12GB RAM. For the exact details corresponding to individual training settings, including the training and test examples used, please refer to the respective figures.

\section{Hyper parameter Analysis}

\begin{figure}[H]{
    \includegraphics[scale=0.21]{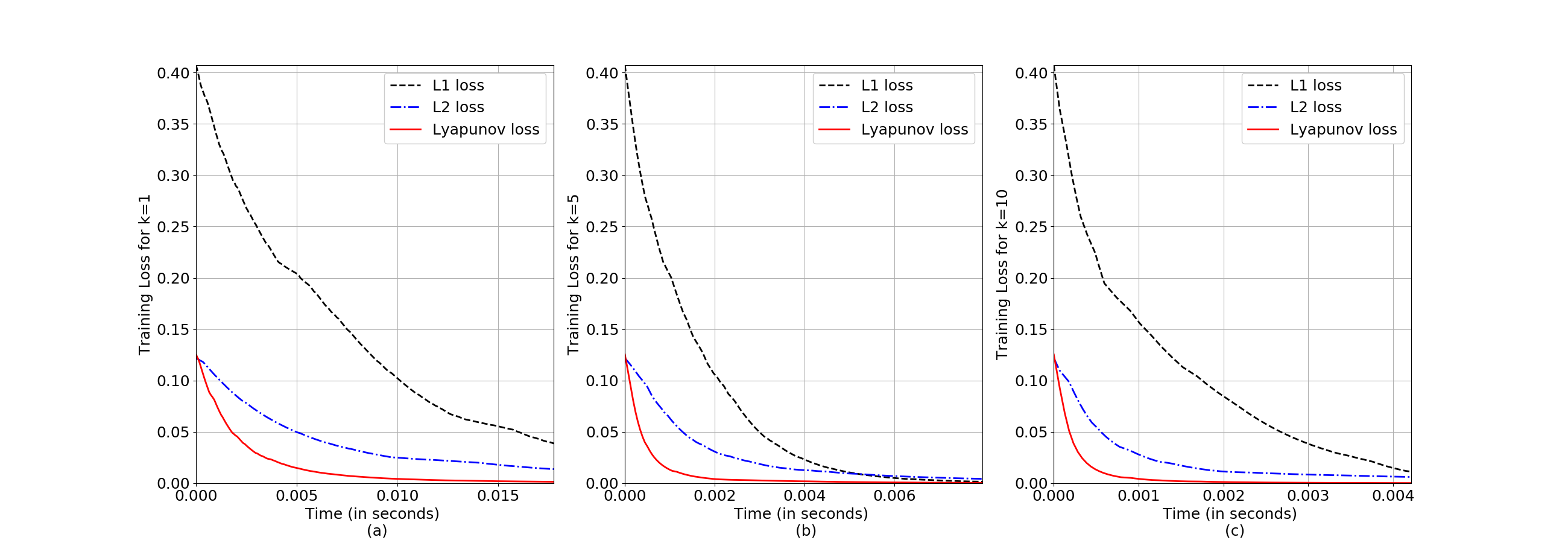}
    \centering
    \caption{Comparison of training convergence with respect to time for different values of tuning parameter, $k$ in a single neuron trained on the Iris dataset. We convert the problem into a binary classification problem by only considering inputs of two of the three classes. All networks are trained for 600 epochs with 80 training examples and 20 test examples.} 
    \label{figure1}}
\end{figure}

\begin{figure}[H]{
    \includegraphics[scale=0.21]{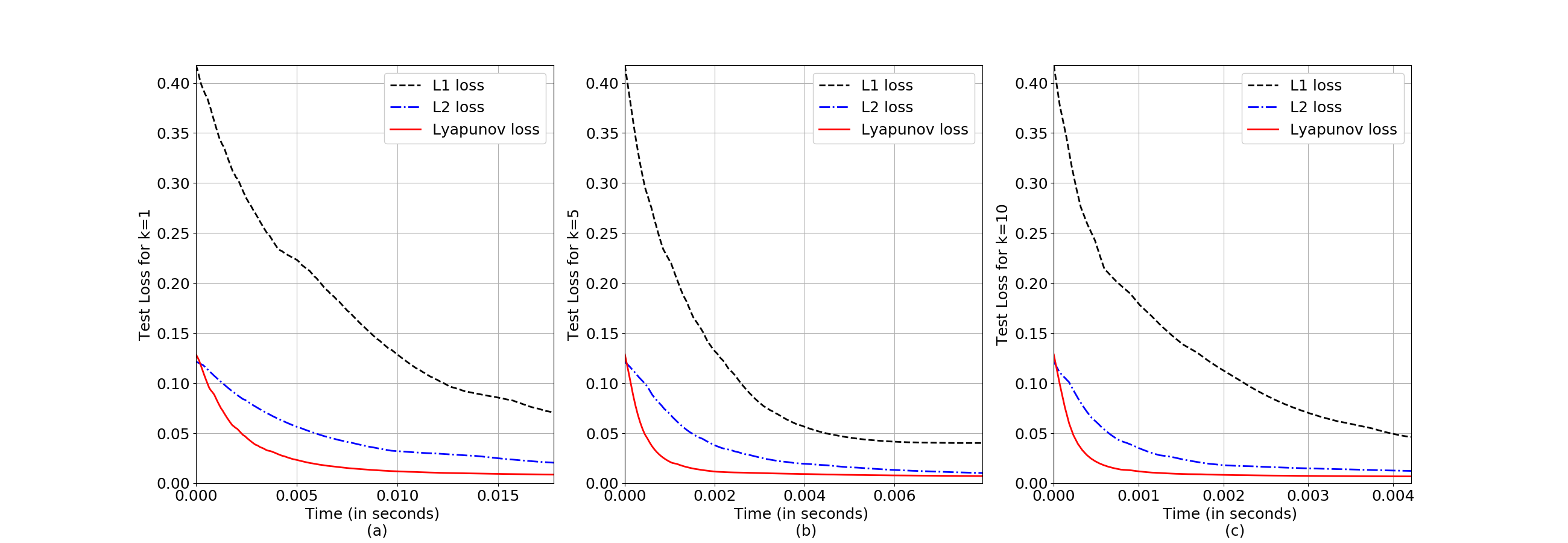}
    \centering
    \caption{Comparison of test loss convergence with respect to time for different values of tuning parameter, $k$ in a single neuron trained on the Iris dataset. This loss is plotted on the results obtained from 20 test examples.} 
    \label{figure2}}
\end{figure}

In this section, we attempt to give an empirical analysis of how changes in the tuning parameter, $k$ affects the training of our neural network. We work with the single neuron case in this discussion for simplicity and ease of understanding. The dataset used for this experiment is the Iris dataset (\citet{Dua:2019}). Existing literature in control theory suggests that a monotonic increase in the tuning parameter, $k$ should result in a monotonic decrease in the corresponding loss. This indicates that increasing $k$ makes the learning more aggressive, especially for the Lyapunov Loss Function. We observe this trend clearly in Figure \ref{figure1}. However, graphs for test loss as shown in Figure \ref{figure2} play an important role here to ensure that the model has accurately learnt features that best represent the input dataset. From Table \ref{settling_time_table}, we observe that the settling time for the training loss of our Lyapunov loss function is always less than $L_1$ and $L_2$, proving that our loss function trains the network aggressively. Though this is a small example, it demonstrates the superiority of our model in terms of settling time when compared to $L_1$ and $L_2$ loss functions. While we work with a fixed value of tuning parameter throughout the training, this can easily be extended to make the tuning gain parameter adaptive. Individual adaptive tuning gains for the multi-layer perceptron will greatly impact the accuracy of the high dimensional model. 

\begin{table}[ht]
  \centering
  \begin{tabular}{cccccc}
    \toprule
    \textbf{Experiment}  & \textbf{Theoretical Upper Bound}  & \multicolumn{3}{c}{\textbf{Experimental Convergence Time}}\\
    \textbf{for single} & (in seconds) & \multicolumn{3}{c}{(in seconds)} \\
    \textbf{neuron case} & & $L_1$ & $L_2$ & Lyapunov\\
    \textbf{on Iris dataset} & & & & Loss function \\
    \hline
    $k=1$ & $\sim20224.176$ & 0.04303 & 0.04305 & 0.01787  \\
    $k=5$ & $\sim4044.83522$ & 0.01158 & 0.01141 & 0.0066  \\
    $k=10$ & $\sim$2022.4176   & 0.00697 & 0.00695 & 0.00463\\ 
    \bottomrule
  \end{tabular}
  \caption{Settling time in seconds for different values of tuning parameter, $k$ for the single neuron case on Iris dataset. We compare the time taken for convergence by three different loss functions, $L_1$, $L_2$ and Lyapunov Loss function. The training conditions were similar for individual cases in the experiment.}
  \label{settling_time_table}
\end{table}

We restrict our discussion to the tuning gain analysis for single neuron as a comprehensive discussion on the same for multi-layer perceptron requires the development of extensive theory to solve the resultant optimal control problem, which is beyond the scope of this work. We hope to address this inverse control problem as future work.

\section{Stability of learning in the presence of amplitude-bounded perturbations}

In this section, we present numerical results for demonstrating robustness of proposed algorithm to bounded perturbations while maintaining convergence of training dynamics. For this case, we perform experiments pertaining to multi-layer perceptrons exclusively due to their ability to give out better representations. We consider training on two separate datasets here: Boston Housing Dataset (\citet{harrison1978hedonic}) and IMDB Wiki Faces Dataset (\citet{Rothe-ICCVW-2015}). While we have solely focused our attention on training convergence in the discussions preceding this section, the trends for test loss become imperative when it comes to any discussion of input perturbations. Ensuring that the model is able to differentiate between the input and its corresponding perturbations and does not end up over-training to fit for the noisy data is important. A good indicator to show that the model is over-training can be when our test loss starts diverging. 

\subsection{Boston Housing Experiment} The Boston Housing Dataset predicts the price of houses in various areas of Boston Mass, given 14 different relevant attributes, like per capita crime rate and pupil-teacher ratio by town. The 506 examples in the data are divided into a 80-20 training-test split to give us 505 training examples and 101 test examples respectively. 

For adding input perturbations, we stick to the premise presented in our theoretical proofs. We asssume that an a priori upper bound, $M$ is known on the amplitude of possible input perturbations. While our algorithm is bounded when it comes to the amplitude, input perturbations with any frequency range are allowed. We add input perturbations to each pixel in the image using a randomized uniform distribution ranging from $(-\Delta x, \Delta x)$. Hence, our additive noise remains in the abovementioned a priori bounds. We consider three cases here, $M=0.1$, $M=0.2$ ,and $M=0.3$. 

Table \ref{settling_time_table_1} presents the theoretical upper bounds for the proposed Lyapunov function's settling time and experimental results for the training conducted for different upper bounds assumed for the input perturbation. We observe that the settling time for the proposed Lyapunov function is similar to the one observed for $L_2$ loss function whereas it performs way better than the $L_1$ loss function. Lyapunov function does show slight delay in converging when compared to $L_2$ loss function. However, when looking at the iterations in which the function converges for $M=0.3$, $L_1$ takes 102021 iterations to converge, $L_2$ function takes 26356 and Lyapunov loss function converges in 24224 iterations. 

\begin{table}[ht]
  \centering
  \begin{tabular}{cccccc}
    \toprule
    \textbf{Experiment}  & \textbf{Theoretical Upper Bound}  & \multicolumn{3}{c}{\textbf{Experimental Convergence Time}}\\
    \textbf{for MLP case} & (in seconds) & \multicolumn{3}{c}{to within 10e-9 (in seconds)} \\
    \textbf{on Boston Housing} & & $L_1$ & $L_2$ & Lyapunov\\
    \textbf{dataset} & & & & Loss function \\
    \hline
    $M=0.1$ & $\sim2.16403992e+13$ & 2341.296 & 674.524 & 896.22 \\
    $M=0.2$ & $\sim5.67913851e+12$ & 2813.108 & 616.958 & 733.191 \\
    $M=0.3$ & $\sim2.59676901e+12$ & 2394.627 & 596.863 & 787.953 \\
    \bottomrule
  \end{tabular}
  \caption{Settling time in seconds for different values of the upper bound on additive input perturbations, $M$ for the multi layer perceptron case on Boston Housing dataset. We compare the time taken for convergence by three different loss functions, $L_1$, $L_2$ and Lyapunov Loss function. The training conditions were similar for individual cases in the experiment.}
  \label{settling_time_table_1}
\end{table}

\subsection{IMDB Wiki Faces Experiment} The IMDB Wiki Faces Dataset has over 0.5 million face images for celebrities with their age and gender labels. Here, we primarily focus on dealing with the age prediction problem for the given face images to differ from the binary classification problem tackled for the single neuron experiments. We also reduce the training dataset to 20,000 images and extract 4,000 images each for validation and test datasets from the remaining images. This example is a better demonstration of how the proposed Lyapunov loss function performs on a larger dataset where there is more data to learn representations from.

We vary the values of $M$ from $(0.1, 0.5)$ with a $0.1$ increment, giving us five training cases. We also add another case where $M = 1.2$. Since we are dealing with normalized inputs in the range of $[0,1]$, this indicates that the additive noise here ends up being more than the input signal itself.  

\begin{figure}[H]{
    \includegraphics[scale=0.7]{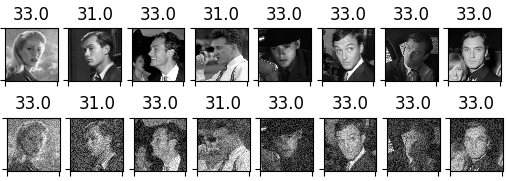}
    \centering
    \caption{Pictorial representation of the effect of random additive noise with an upper bound of $0.2$ on IMDB Wiki Faces Dataset. In this experiment, we work with monochrome images for the sake of simplicity.} 
    \label{figure3}}
\end{figure}

Figure \ref{figure3} shows us how the noise affects the training images. We take the specific case where $M=0.2$. We can clearly observe that the images obtained after adding input perturbations happens to be quite noisy. Since we also go beyond $M=0.2$ for our analysis, we can safely conclude that the analysis done for input perturbations is sufficiently good for majority of the cases. In fact, we also consider a scenario where $M=1.2$. That is, the noise is more in amplitude when compared to our input signal. Although this might not be observed in real world images, this experiment is quite relevant for astronomical or physics based data. 

\begin{figure}[H]{
    \includegraphics[scale=0.08]{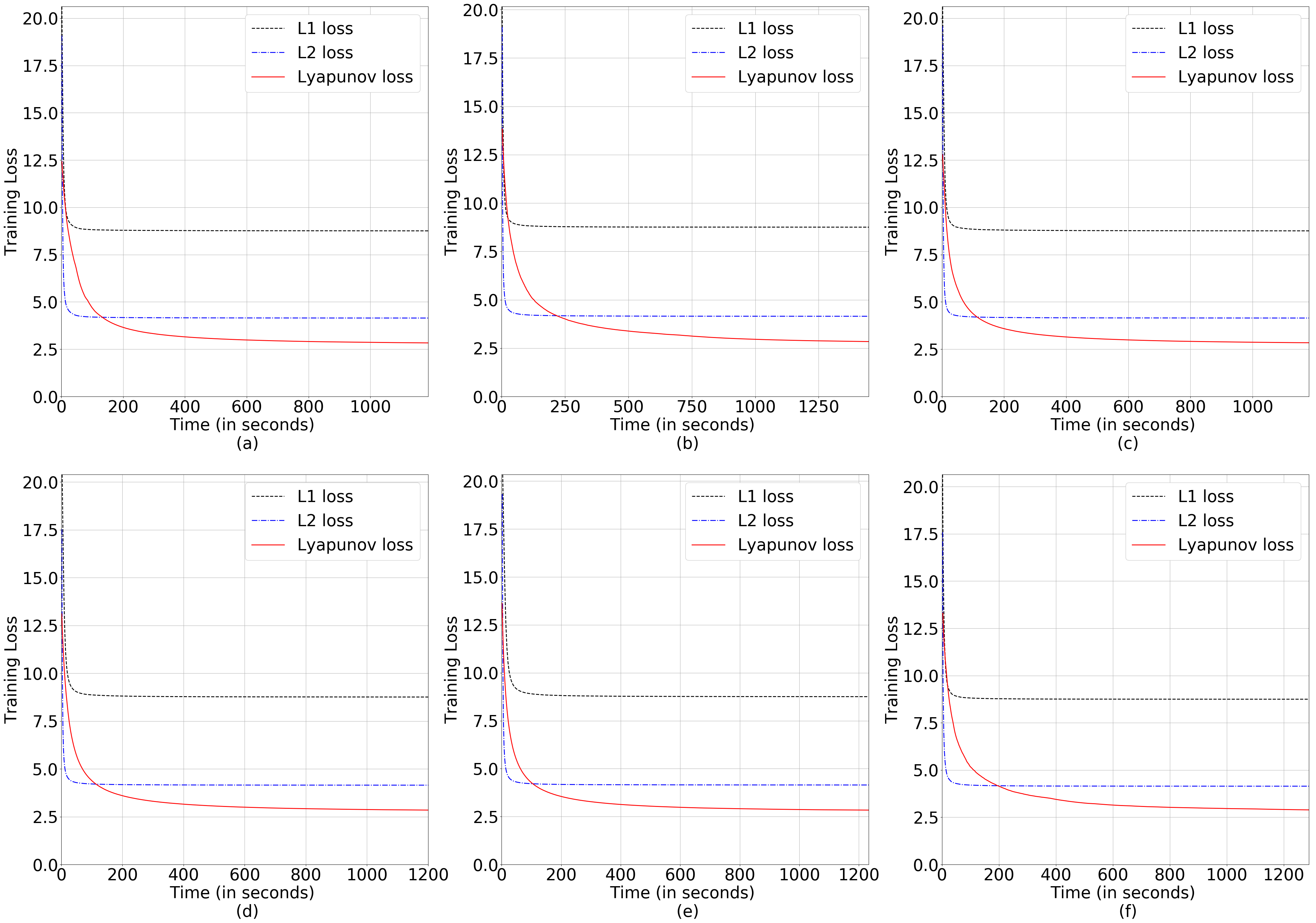}
    \centering
    \caption{Comparison of training loss convergence with respect to time for different values of input perturbations, $\Delta x$ for a multi-layer perceptron trained on the IMDB Wiki Faces dataset. The a priori upper bound on input perturbations are as follows: (a) $\Delta x = 0.1$, (b) $\Delta x = 0.2$, (c)$\Delta x = 0.3$, (d) $\Delta x = 0.4$, (e) $\Delta x = 0.5$, and (f) $\Delta x = 1.2$.} 
    \label{figure4}}
\end{figure}

Figure \ref{figure4} shows that the proposed loss function still manages to achieve steady state error that is lower than $L_1$ and $L_2$ loss functions. Since we are dealing with a more complex network with higher dimensional data and more control parameter to optimize, the error values will take longer to converge to zero. Looking closely, we observe that the presence of noise does delay the convergence since the network needs to learn the difference between the noise and input data in addition to learning the relation between the faces and apparent age of the celebrities. 

\begin{figure}[H]{
    \includegraphics[scale=0.08]{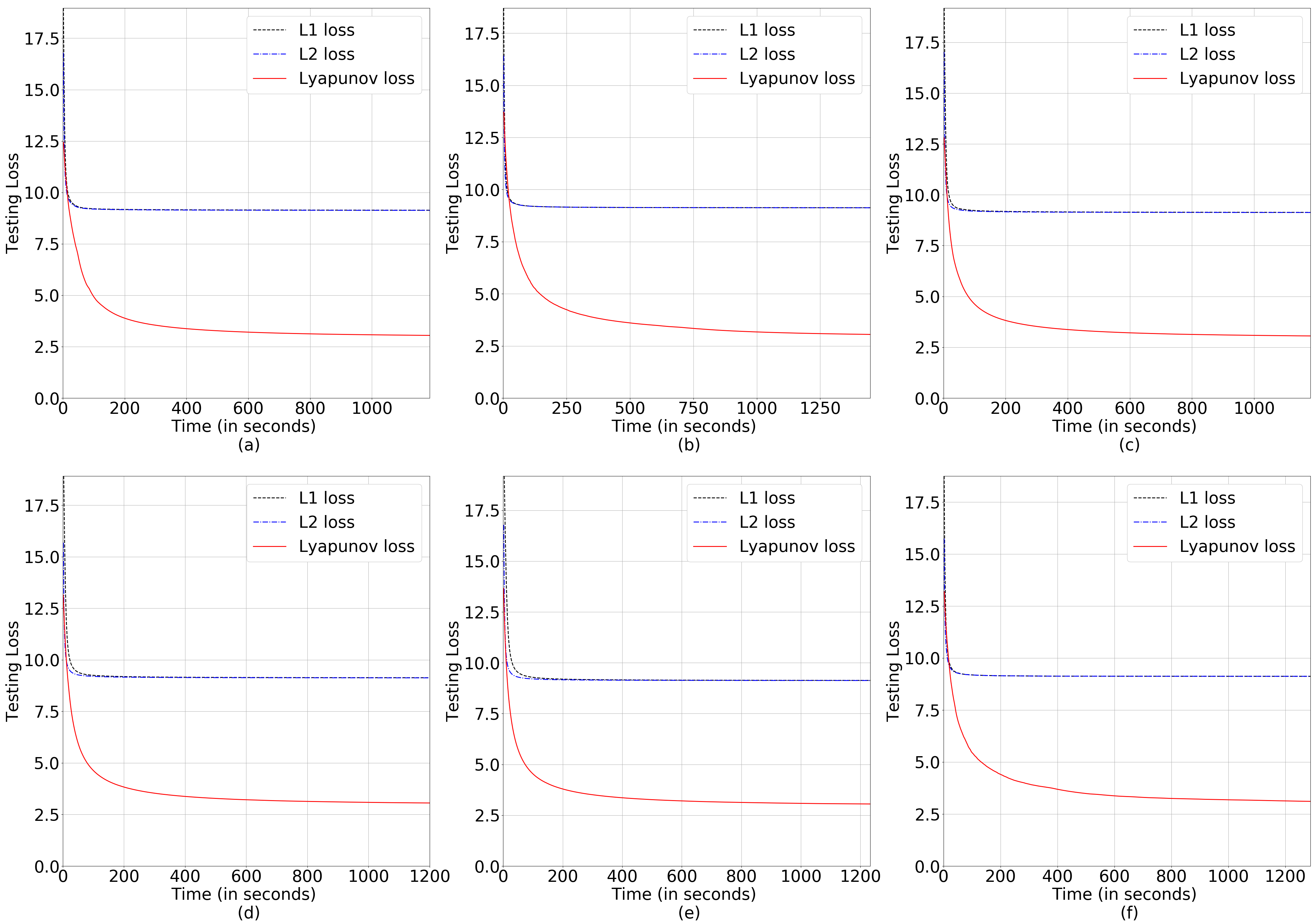}
    \centering
    \caption{Comparison of test loss convergence with respect to time for different values of input perturbations, $\Delta x$ for a multi-layer perceptron trained on the IMDB Wiki Faces dataset.The a priori upper bound on input perturbations are as follows: (a) $\Delta x = 0.1$, (b) $\Delta x = 0.2$, (c)$\Delta x = 0.3$, (d) $\Delta x = 0.4$, (e) $\Delta x = 0.5$, and (f) $\Delta x = 1.2$.} 
    \label{figure5}}
\end{figure}

While Figure \ref{figure4} follows the trends observed in the previous results, what is interesting is the test convergence results shown in Figure \ref{figure5}. It is noted that the proposed Lyapunov function ends up having half the loss that $L_1$ and $L_2$ show for this example. This is promising as it also demonstrates that the test loss mimics the training loss quite closely, only proving to be slightly worse for our test cases. The discrepancy observed between the training and test loss for $L_2$ loss function when compared to that of the Lyapunov Loss function shows that that our loss function performs better when it comes to handling noise and is robust in its learned representations. We would like to point out that the above results are obtained in the absence of any hyper parameter tuning or optimization. With better hyper parameter tuning and optimization, we expect the model to perform better and reduce the observed loss further.

\bibliography{supplement}
\bibliographystyle{plainnat}